\documentclass[10pt,twocolumn,letterpaper]{article}

\usepackage{iccv}
\usepackage{times}
\usepackage{epsfig}
\usepackage{graphicx}
\usepackage{amsmath}
\usepackage{amssymb}
\usepackage{authblk}

\usepackage[breaklinks=true,bookmarks=false]{hyperref}

\iccvfinalcopy 

\def\iccvPaperID{4533} 

\ificcvfinal\fi

\begin{document}

\title{Continual Learning by Asymmetric Loss Approximation \\ with Single-Side Overestimation}
%

\author{
Dongmin Park$^{1,2}$\hspace{1.0cm} Seokil Hong$^1$ \hspace{1.0cm} Bohyung Han$^{1}$\hspace{1.0cm} Kyoung Mu Lee$^1$\\
\hspace{-0.7cm}
$^1$ECE \& ASRI, Seoul National University, Korea \hspace{1.0cm}
$^2$Samsung Electronics, Korea \\
{\small \texttt {dmpark04@gmail.com, \{hongceo96, bhhan, kyoungmu\}@snu.ac.kr}}
}

\maketitle
\ificcvfinal\thispagestyle{empty}\fi

\begin{abstract}
Catastrophic forgetting is a critical challenge in training deep neural networks.
Although continual learning has been investigated as a countermeasure to the problem, it often suffers from the requirements of additional network components and the limited scalability to a large number of tasks.
We propose a novel approach to continual learning by approximating a true loss function using an asymmetric quadratic function with one of its sides overestimated.
Our algorithm is motivated by the empirical observation that the network parameter updates affect the target loss functions asymmetrically.
In the proposed continual learning framework, we estimate an asymmetric loss function for the tasks considered in the past through a proper overestimation of its unobserved sides in training new tasks, while deriving the accurate model parameter for the observable sides.
In contrast to existing approaches, our method is free from the side effects and achieves the
state-of-the-art accuracy that is even close to the upper-bound performance on several challenging benchmark datasets.
\end{abstract}


\section{Introduction}
\label{sec:introduction}
It is common to learn machine learning models versatile for multiple tasks in an incremental manner when new tasks are given one by one, not in a batch.
Continual learning is a concept to learn a model for a large number of tasks sequentially without forgetting knowledge obtained from the preceding tasks, where the data in the old tasks are not available any more during training new ones.

Catastrophic forgetting is a critical issue in realizing continual learning using deep neural networks. 
With a na\"ive stochastic gradient descent (SGD) method, deep neural networks easily forget the knowledge obtained from the earlier tasks while adapting to the new information quickly from the incoming tasks~\cite{McCloskey1989,French1999}.
This is mainly because, without any countermeasure, the optimization based on the losses of new tasks may not be desirable to retain the knowledge about the tasks learned in the past.
The catastrophic forgetting problem limits the capability and potential of deep neural networks to be applied to dynamic real-world problems that require continuous adaptation to new environments.
\begin{figure}[t]
	\begin{center}
		\includegraphics[width=1\linewidth]{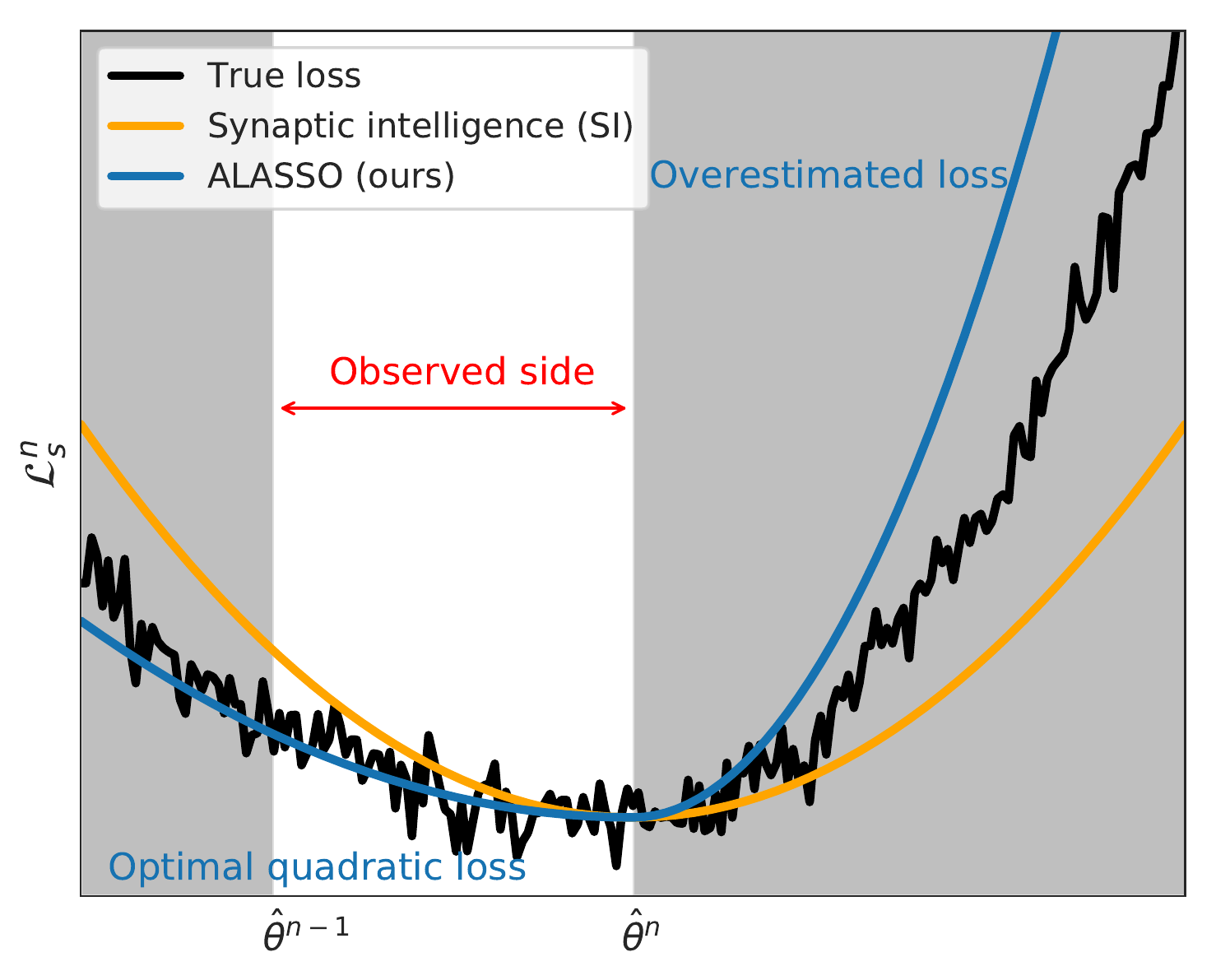}
	\end{center}
	\vspace{-0.3cm}
	\caption{Conceptual diagram to illustrate why our loss approximation is required. The symmetric loss approximations such as the quadratic approximation used in SI \cite{OnlineEWC}(yellow) is prone to underestimate the unobserved sides of potentially asymmetric real surrogate loss functions (black). We claim that properly introduced asymmetry in the loss approximation (blue) prevents this problem.
		Note that the other side are observed during the optimization of the network in training the previous tasks and we can derive a more accurate parameter for the quadratic optimization of the surrogate function. }
	\vspace{-0.5cm}
	\label{fig:ConceptualDiagramExplainingAsymmetricLossApproximation}
\end{figure}
The continual learning or life-long learning is a well-known framework to handle the catastrophic forgetting problem.
It can be categorized into three categories: architectural approach, functional approach, and structural regularization approach.
The architectural and functional approaches typically need additional network components and/or batch processing.
The structural regularization methods work well for a limited number of tasks, but often have scalability issues to many tasks.

This paper presents a novel continual learning framework based on asymmetric loss approximation with single-side overestimation (ALASSO), which effectively adapts to a large number of tasks.
ALASSO approximates the true loss functions corresponding to the previously considered tasks asymmetrically by overestimating their unobserved sides in the parameter space while deriving the accurate quadratic approximation on the observed sides.
Figure~\ref{fig:ConceptualDiagramExplainingAsymmetricLossApproximation} illustrates the main concept of our approach; it computes the optimal parameter through a quadratic approximation in the observed side (left) while using a steep surrogate quadratic function in the unobservable side (right).
The proposed algorithm also decouples the hyperparameters for the current surrogate loss approximation and the surrogate loss change of the previous tasks.
This approach is motivated by our observation that updating the model parameters of deep neural networks affects target losses asymmetrically and that using the overestimated loss functions is relatively safe for the optimization without the training data of the previous tasks.

In contrast to the existing approaches, the proposed technique is free from the additional memory requirement to store the information about the previous tasks and the overhead of batch processing.
ALASSO achieves the state-of-the-art performance and is even close to the accuracy upper-bounds in several challenging benchmark datasets, including the permuted MNIST, the split CIFAR-10/CIFAR-100 and the split Tiny ImageNet.
In particular, we demonstrate promising results in terms of scalability and robustness to a large number of tasks.

The contribution of our work is summarized as follows:
\begin{itemize} 
\item We propose a novel continual learning framework by overestimating the unobserved side of a loss function in the current task and approximating the loss using an asymmetric quadratic function. This strategy facilitates a reliable loss estimation even without the training data of the previous tasks.
\item Our algorithm provides an accurate solution of the loss approximation for the previous tasks, which allows to derive the best approximation of a quadratic surrogate loss function on the observed side.
\item The proposed technique achieves the outstanding performance on several challenging benchmark datasets by large margins. 
\end{itemize}

The rest of the paper is organized as follows.
We first discuss the related works in Section~\ref{sec:related}, and present the technical details of our algorithm, ALASSO, in Section~\ref{sec:proposed}.
Section~\ref{sec:experiment} demonstrates the experimental results with analysis and Section~\ref{sec:conclusion} concludes this paper.

\section{Related Work}
\label{sec:related}

Continual learning algorithms are categorized into three groups~\cite{OnlineEWC}: architectural, functional, and structural regularization approaches.
This section discusses the existing methods in individual categories and and their characteristics briefly.

\subsection{Architectural Approaches}
The approaches in this category realize continual learning by freezing model parameters learned from the previous tasks and/or providing limited architectural variations for learning the new tasks~\cite{ProgressiveNeuralNetworks,Lee2016,ExpertGate,DEN}.
This framework often requires additional network components for the new tasks and the size of the network gradually increases in principle.
This drawback limits the applicability to large-scale problems and is not appropriate for the configurations with limited resources such as embedded systems.
Existing methods~\cite{GradientEpisodicMemory,DeepGenerativeReplay,FearNet,MemoryBasedParameterAdaptation,hu2018overcoming} often keep track of the data in the previous tasks by allocating additional memory space called episodic memory or by generating training examples using generative adversarial networks~\cite{GAN}, but they suffer from substantial memory demands to store the information of the previous tasks.
Some approaches in this category aim to improve performance using different nonlinearities such as ReLU, MaxOut, and local winner-take-all~\cite{CompeteToCompute,Goodfellow2013}.

An interesting work among the architectural approaches, instead of adding new network components, employs a network compression technique to identify unused or rarely used parts in the target network and allows the free space to store the information of new tasks~\cite{PackNet,Conceptor}.
It alleviates the drawbacks of the standard approaches discussed earlier, but requires the substantial overhead of batch processing for network compression.

\subsection{Functional Approaches}
The functional approaches~\cite{LWF,LessForgetfulLearning,progresscompress,icarl} often incorporate knowledge distillation to establish continual learning.
The previously learned networks are fixed and used for feature computation in training new tasks.
The new networks are encouraged to learn the representations that are coherent to the features computed in the previous networks.
However, the feature coherency of new examples with respect to the old ones does not always ensure the output similarity of the examples in the past and current tasks.
Existing functional approaches are required to store and evaluate the previous networks for training the model for the new task, which incurs additional computational overhead.

\subsection{Structural Regularization Approaches}
The structural regularization approaches typically augment a penalty term to the original loss functions and discourage the updates of parameters critical to the previously learned tasks.
Elastic weight consolidation (EWC)~\cite{EWC} and synaptic intelligence (SI)~\cite{OnlineEWC} employ a surrogate quadratic loss as an approximation of the real loss functions of the previous tasks.
Although this is simple and effective for a small number of tasks, its performance drops drastically when there are a larger number of tasks.
Memory aware synapses (MAS)~\cite{MAS} estimates the importance of the weights in an
unsupervised manner.
Incremental moment matching (IMM)~\cite{IMM} additionally performs a separate model-merging step after learning a new task.
Variational continual learning (VCL)~\cite{VCL} combines the online variational inference~\cite{Ghahramani2000,Sato2001,Broderick2013} with Monte Carlo sampling~\cite{Blundell2015} using neural networks, but it requires a relatively large amount of computational cost to infer an approximate posterior distribution.
On the other hand, \cite{HAT} proposes a task-specific hard attention mechanism, but, as in \cite{ProgressiveNeuralNetworks,PathNet,LWF}, the requirement of multi-head outputs---a separate output per task---limits the number of target domains considered concurrently.

Our framework falls in the structural regularization category.
We design a new loss function appropriate for continual learning while it does not incur extra computational overhead such as additional network components and the need for occasional batch processing.

\begin{figure}[t]
	\begin{center}
		\includegraphics[width=0.9\linewidth]{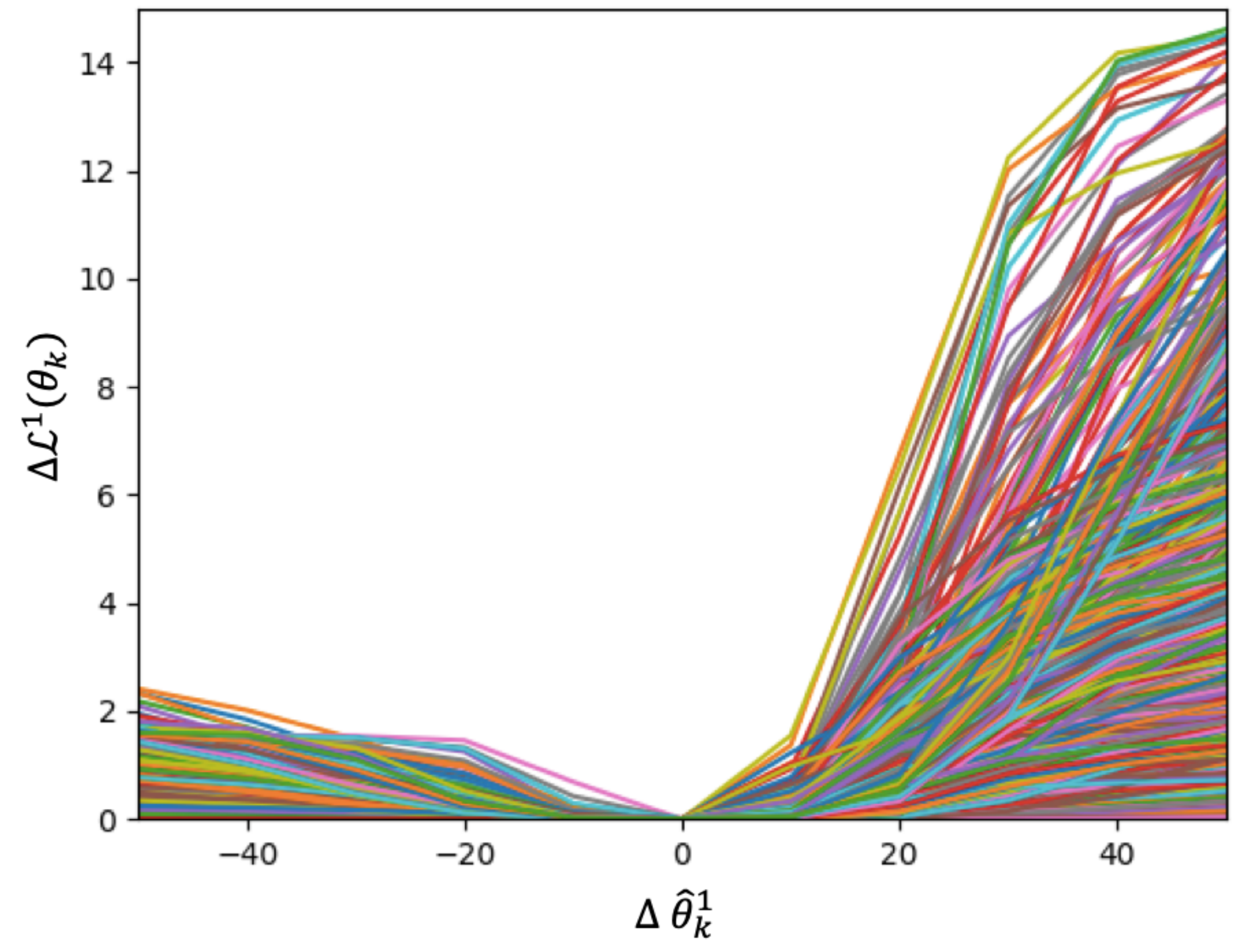}
	\end{center}
	\vspace{-0.2cm}
	\caption{Illustration of asymmetric characteristics of a loss function in terms of each model parameter $\hat{\theta}_k$, where $k$ is the index of model parameter.  Each graph shows how much loss $\mathcal{L}^1$ changes by updating each parameter $\hat{\theta}_k^1$ after learning the first task in the permuted MNIST dataset.}
	\vspace{-0.2cm}
	\label{fig:WeightChangeAnalysis}
\end{figure}

\section{Synaptic Intelligence (SI)}
\label{sec:preliminaries}

The proposed method, referred to as ALASSO, is closely related to synaptic intelligence (SI)~\cite{OnlineEWC}.
We first discuss the main idea of SI and then point out its critical problems before presenting our novel idea.

\subsection{Quadratic Approximation of Loss}

SI is categorized as a structural regularization method, which employs a static network architecture and does not use additional memory throughout continual learning process.
To prevent the catastrophic forgetting problem and maintain the performance with respect to the previous tasks while adapting to a new task $n$, this technique introduces a surrogate loss function $\mathcal{L}_s^{n-1}$, which approximates the loss of the previous tasks and plays a role as a regularizer.
Assuming that the surrogate loss is a quadratic function, the total loss function to learn the $n^\text{th}$ task, $\tilde{\mathcal{L}}^n$, in terms of the model parameter $\theta_k$ is approximated by
\begin{equation} \label{eq1}
\begin{split}
\tilde{\mathcal{L}}^n &= \mathcal{L}^n +  c \hspace{0.05cm} \mathcal{L}_s^{n-1} \\
&= \mathcal{L}^n + \underbrace{ c \sum_{k} \hat{\Omega}_{k}^{n-1}\left(\theta_k - \hat{\theta}_k^{n-1} \right)^2}_{\text{surrogate loss}},
\end{split}
\end{equation}
where $\mathcal{L}^n$ denotes a loss for the current task $n$, $\hat{\theta}_k^{n-1}$ is the weight in the $k^\text{th}$ dimension of the estimated parameter for the previous tasks until $n-1$, $\hat{\Omega}^{n-1}_k$ is a coefficient for the corresponding model parameter, and $c$ is a hyperparameter for the surrogate loss.

We minimize the total loss, $\tilde{\mathcal{L}}^n$, with respect to $\theta_k$, and obtain the optimized parameters, $\hat{\theta}_k^{n}$.
Assuming that the surrogate loss function for each parameter up to the $n^\text{th}$ task is defined by a quadratic function as
\begin{equation}
\mathcal{L}_s^n(\theta_k) = \hat{\Omega}^{n}_k\left(\theta_k - \hat{\theta}^{n}_k \right)^2,
\label{eq:quadratic_surrogate}
\end{equation}  
$\hat{\Omega}^n_k$ is derived approximately by 
\begin{equation} \label{eq2}
\hat{\Omega}^n_k \approx \frac{\omega^n_k}{\left(\hat{\theta}_k^n - \hat{\theta}_k^{n-1}\right)^2} + \hat{\Omega}^{n-1}_k,
\end{equation}
where $\omega^n_k$ denotes the difference between the losses before and after training a new task $n$, \textit{i.e.}, $\omega_k^n = \mathcal{L}^n(\hat{\theta}_k^{n-1}) - \mathcal{L}^n(\hat{\theta}_k^{n})$.
Refer to \cite{OnlineEWC} for more details about SI.

\subsection{Underestimation of Loss}

SI employs a symmetric loss function for approximation with respect to the previous tasks while it can observe only a single side of the symmetric function along the trajectory of parameter update from $\hat{\theta}_k^{n-1}$ to $\hat{\theta}_k^n$ during the optimization process for the $n^\text{th}$ task.
In other words, the surrogate loss function is assumed to be symmetric as shown in Eq.~\eqref{eq1} and may not be able to model its unobserved side accurately unless the true loss function is symmetric.

However, it turns out that the true loss functions are typically asymmetric and the symmetric loss functions adopted by SI are prone to underestimate the true losses in practice.
Figure~\ref{fig:WeightChangeAnalysis} illustrates the variation of the losses with respect to the changes of each model parameter; $x$-axis denotes the offset from the optimal model parameter and $y$-axis is the difference of the cross-entropy losses, which are obtained from the first task of the permuted MNIST dataset.
Note that, to visualize the asymmetry more clearly, the steeper halves of the individual graphs are located on the right hand side.
Based on this observation, we consider an asymmetric loss function formulation for a more accurate and stable estimation of the surrogate loss function parametrized by $\hat{\Omega}^n_k$. 


\section{Proposed Algorithm}
\label{sec:proposed}
We propose a novel structure regularizer, which mitigates the limitations of SI.
The main contributions of our algorithm, ALASSO, include the introduction of the asymmetric loss function with single-side overestimation and the more accurate quadratic approximation of the loss function, which lead to remarkable performance improvement. 
This section presents the details about ALASSO, especially, in comparison to the existing approach, SI.

\subsection{Overview}
\label{sub:overview}
Our algorithm overestimates the unobserved sides of the approximate loss function and allows the models to learn under a harsher condition.
It also derives the accurate parameter estimation of the approximate quadratic loss functions on their observed sides.
To accelerate the optimization procedure and handle the conflicts between the loss computation of the current task and the loss approximation of the previous tasks, we introduce a hyperparameter decoupling technique although the values of the decoupled hyperparameters should be identical conceptually. 
The proposed algorithm inherits the merits of the standard structural regularization approaches  such as online learning and local updates while providing the capability to maintain the crucial knowledge about the prior tasks by making the models less adaptive without the observation of the loss function.
We now discuss the technical contributions and characteristics of our continual learning framework, ALASSO.

\subsection{Asymmetric Loss Approximation}
\label{sub:asymmetric}

One possible option for a better structural regularizer in continual learning is asymmetric loss approximation of the previous tasks.
We believe that the symmetric regularizer as in SI overly simplifies the true loss functions and have a critical limitation in maintaining the knowledge obtained from the previous tasks.
Figure~\ref{fig:ConceptualDiagramExplainingAsymmetricLossApproximation} illustrates why asymmetric loss approximation is effective. 
The approximate quadratic loss functions may be sufficiently accurate on the sides, where the true losses are observable along the model parameter updates during training.
However, they may incur substantial error on their unobserved sides, so it is dangerous to assume that the true loss functions are symmetric, which is supported by Figure~\ref{fig:WeightChangeAnalysis}.

Based on this motivation, we now propose a simple but effective approximation approach of the true loss functions.
We believe that the coefficient, $\hat{\Omega}_k^n$, is reliable if the learnable parameter ${\theta}_k$ and the fixed parameter $\hat{\theta}^{n-1}_k$ are on the same side from $\hat{\theta}^{n}_k$ because ${\theta}_k$ can observe the true loss function during optimization process. 
Conversely, if $\theta_k$ and $\hat{\theta}^{n-1}_k$ are on the opposite sides with respect to $\hat{\theta}^{n}_k$, $\hat{\Omega}_k^n$ is unreliable.
Figure~\ref{fig:VisualizeAsymmetricQuadraticLoss} visualizes the relations between the parameters for the both cases.

To model the relationship between the variables, we introduce the following function:
\begin{figure}[t]
	\begin{center}
		\includegraphics[width=0.9\linewidth]{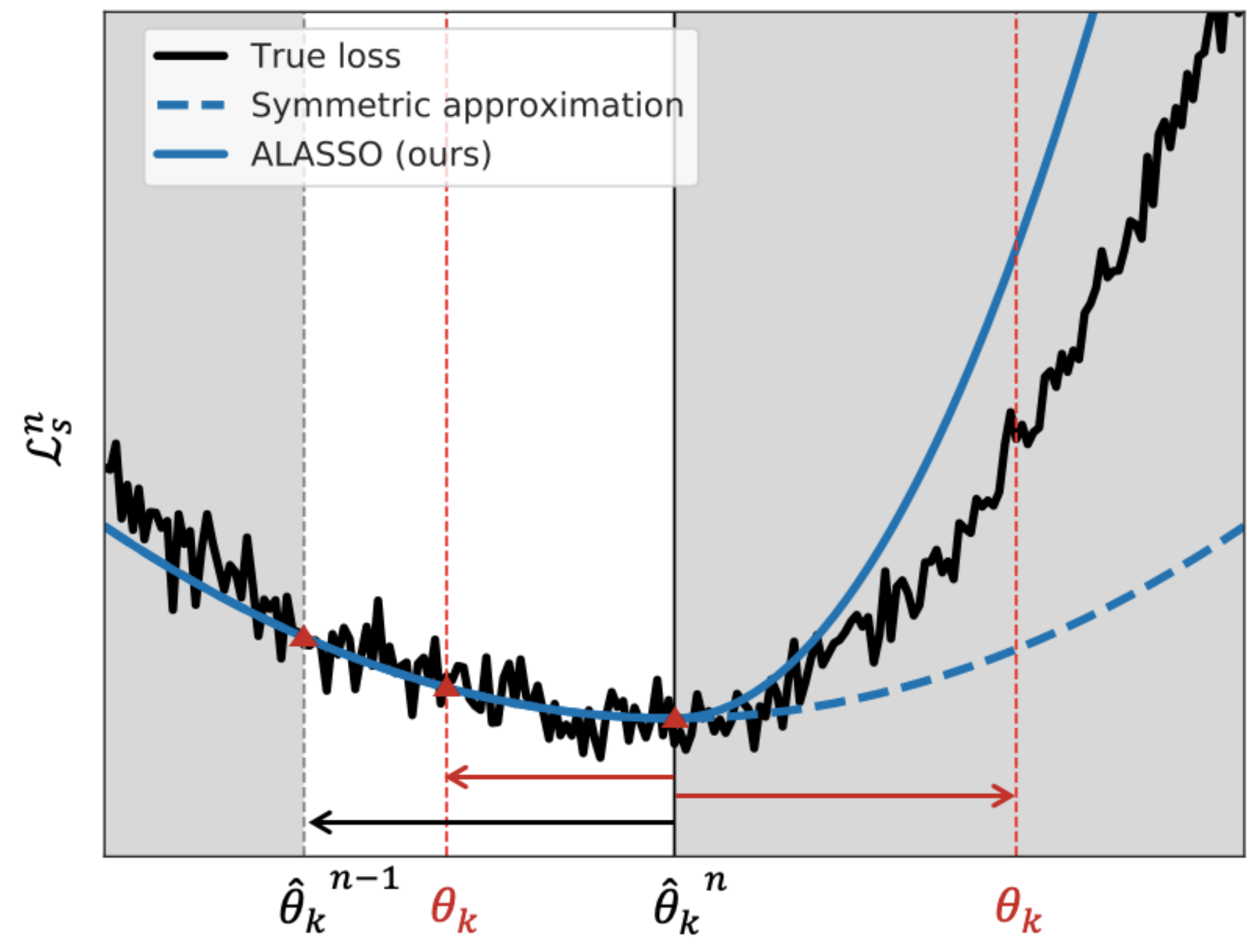}
		\vspace{-0.1cm}
	\end{center}\caption{Visualization of quadratic surrogate loss functions based on the two piece-wise approximation of the true loss. When learning the $n^\text{th}$ task, $\theta_k$ moves by a gradient descent method and the loss approximation is performed differently depending on the location of $\theta_k$---whether it is on the observed side or not. Note that the area between $\hat{\theta}_k^{n-1}$ and  $\hat{\theta}_k^{n} $ is observed.
	}
	\vspace{-0.1cm}
	\label{fig:VisualizeAsymmetricQuadraticLoss}
\end{figure}
\begin{equation}
\alpha(\theta_k) \equiv \left( \theta_k-\hat{\theta}_k^n \right) \left( \hat{\theta}_k^{n-1}-\hat{\theta}_k^n \right).
\label{eq:sign}
\end{equation}
In our framework, since the true loss function is not fully observable, we introduce an additional parameter, $\hat{\Omega}_k^n (\geq 0)$, for the quadratic approximation, which results in the asymmetric loss function as
%
\begin{equation}
\hspace{-0.15cm} \mathcal{L}_s^n \left( \theta_k, a \right) \hspace{-0.05cm} = \hspace{-0.05cm} \begin{cases}
\hat{\Omega}_k^n \left(\theta_k-\hat{\theta}_k\right)^2&\text{if $\alpha(\theta_k)  > 0$ }\\
\left(a\hat{\Omega}_k^n+\epsilon\right)  \hspace{-0.1cm}  \left(\theta_k-\hat{\theta}_k\right)^2&\text{if $\alpha(\theta_k) \leq 0 $ }
\end{cases} \hspace{-0.1cm} ,
\label{eq:asymmetric_loss}
\end{equation}
where $a (>1)$ is a hyperparameter to control the degree of overestimation and $\epsilon$ is a small positive number to make the loss overestimated even when $\hat{\Omega}_k^n = 0$\footnote{Since $\hat{\Omega}_k^n$ is non-negative, $(a \hat{\Omega}_k^n+\epsilon) > \hat{\Omega}_k^n$.}
From now, we omit the hyperparameter $a$ in $\mathcal{L}_s^n \left( \theta_k, a \right)$ for notational simplicity, \textit{i.e.}, $\mathcal{L}_s^n \left( \theta_k, a \right) \equiv \mathcal{L}_s^n \left( \theta_k \right)$

The asymmetric loss function in Eq.~\eqref{eq:asymmetric_loss} is used to define the total loss $\tilde{\mathcal{L}}^n$, which is given by
\begin{align}
\label{eq:oala_surrogate}
\tilde{\mathcal{L}}^n &= \mathcal{L}^n +  c \mathcal{L}_s^{n-1} \\
&= \mathcal{L}^n +  c \sum_k \mathcal{L}_s^{n-1}(\theta_k). \nonumber
\end{align}
Note that $\mathcal{L}_s^{n-1}$ can also be interpreted as a regularizer for the $n^\text{th}$ task.
One remaining concern is how to compute $\hat{\Omega}_k^n$ in Eq.~\eqref{eq:asymmetric_loss}, which is discussed next.

\subsection{Accurate Quadratic Approximation }
\label{sub:quadratic}

In addition to the overestimation of the true loss in the unobserved sides for a new task $n$, our algorithm estimates the optimal coefficient $\hat{\Omega}_k^n$ for the quadratic approximation of the loss function in its observed sides.
Note that, as presented in Eq.~\eqref{eq:asymmetric_loss}, $\hat{\Omega}_k^n$ affects the approximation of the loss function in both sides; it is critical to derive $\hat{\Omega}_k^n$ accurately for performance improvement.

The quadratic surrogate loss function in SI is determined by approximating its parameters $\hat{\Omega}_k^n$ as in Eq.~\eqref{eq2}, which is derived from Eq.~\eqref{eq:quadratic_surrogate}.
Instead of using the equation, we present a new derivation that leads to the exact quadratic approximation, which is given by
\begin{align} 
\hat{\Omega}^n_k &= \frac{\mathcal{L}_s^n\left(\theta_k\right)}{\left(\theta_k - \hat{\theta}_k^n\right)^2} = \frac{\mathcal{L}_s^n \left(\hat{\theta}_k^{n-1}\right)}{\left(\hat{\theta}_k^{n-1} - \hat{\theta}_k^n \right)^2} \nonumber \\
&= \frac{\mathcal{L}_s^{n}\left(\hat{\theta}_k^{n-1}\right)- \mathcal{L}_s^{n}\left(\hat{\theta}_k^{n}\right)}{\left(\hat{\theta}_k^{n-1} - \hat{\theta}_k^n \right)^2} \nonumber \\
&= \frac{\omega^n_k + \omega^{1:(n-1)}_k}{\left(\hat{\theta}_k^n -\hat{\theta}_k^{n-1}\right)^2}.
\label{eq:Omega}
\end{align}
Note that $\hat{\Omega}_k^n$ means the change of the surrogate loss $\mathcal{L}^n_s$ when the model parameter changes from $\hat{\theta}^{n-1}_k$ to $\hat{\theta}^{n}_k$ during the learning process of the $n^\text{th}$ task.
The following properties and definitions are required to derive Eq.~\eqref{eq:Omega}:
\begin{align}
\mathcal{L}_s^{n}\left(\hat{\theta}_k^{n}\right) &\propto \left( \hat{\theta}^{n}_k - \hat{\theta}^{n}_k \right)^2 = 0 \\
\mathcal{L}_s^{n}\left( \hat{\theta}_k^{n} \right) &= \mathcal{L}^{n}\left( \hat{\theta}_k^{n} \right) + c\mathcal{L}_s^{n-1}\left( \hat{\theta}_k^{n} \right)
\end{align}
and
\begin{align}
\label{eq:curr_omega}
\omega_k^n &\equiv \mathcal{L}^n \left( \hat{\theta}_k^{n-1} \right) - \mathcal{L}^n \left( \hat{\theta}_k^n \right) \\
\omega^{1:(n-1)}_k &\equiv c \left( \mathcal{L}_s^{n-1} \left( \hat{\theta}_k^{n-1} \right) - \mathcal{L}_s^{n-1} \left( \hat{\theta}_k^{n} \right) \right) \nonumber \\
&= -c \hspace{0.05cm} \mathcal{L}_s^{n-1}\left(\hat{\theta}_k^{n}\right).
\label{eq:past_omega}
\end{align}

The second equality of Eq.~\eqref{eq:Omega} is simply obtained from plugging $\hat{\theta}_{k}^{n-1}$ into $\theta_k$ in both the numerator and denominator.
The numerator in Eq.~\eqref{eq:Omega} has two terms corresponding to Eq.~\eqref{eq:curr_omega} and \eqref{eq:past_omega}, which implies that $\hat{\Omega}_k^n$ is given by the sum of the two parameter importances---the loss differences with respect to the parameter changes---of the new and previous tasks.
Eq.~\eqref{eq:Omega} is actually different from Eq.~\eqref{eq2} only in the importance term of the previous tasks.
In principle, the parameter importance of the previous tasks should be updated depending on the identified local minimum ($\hat{\theta}_k^n$) of the quadratic function.
Note that our formulation presented in Eq.~\eqref{eq:Omega} realizes this exactly while Eq.~\eqref{eq2} uses the fixed importance, $\hat{\Omega}_k^{n-1}$, resulting in poor approximation.

\begin{figure}[t]
	\begin{center}
		\includegraphics[width=0.95\linewidth]{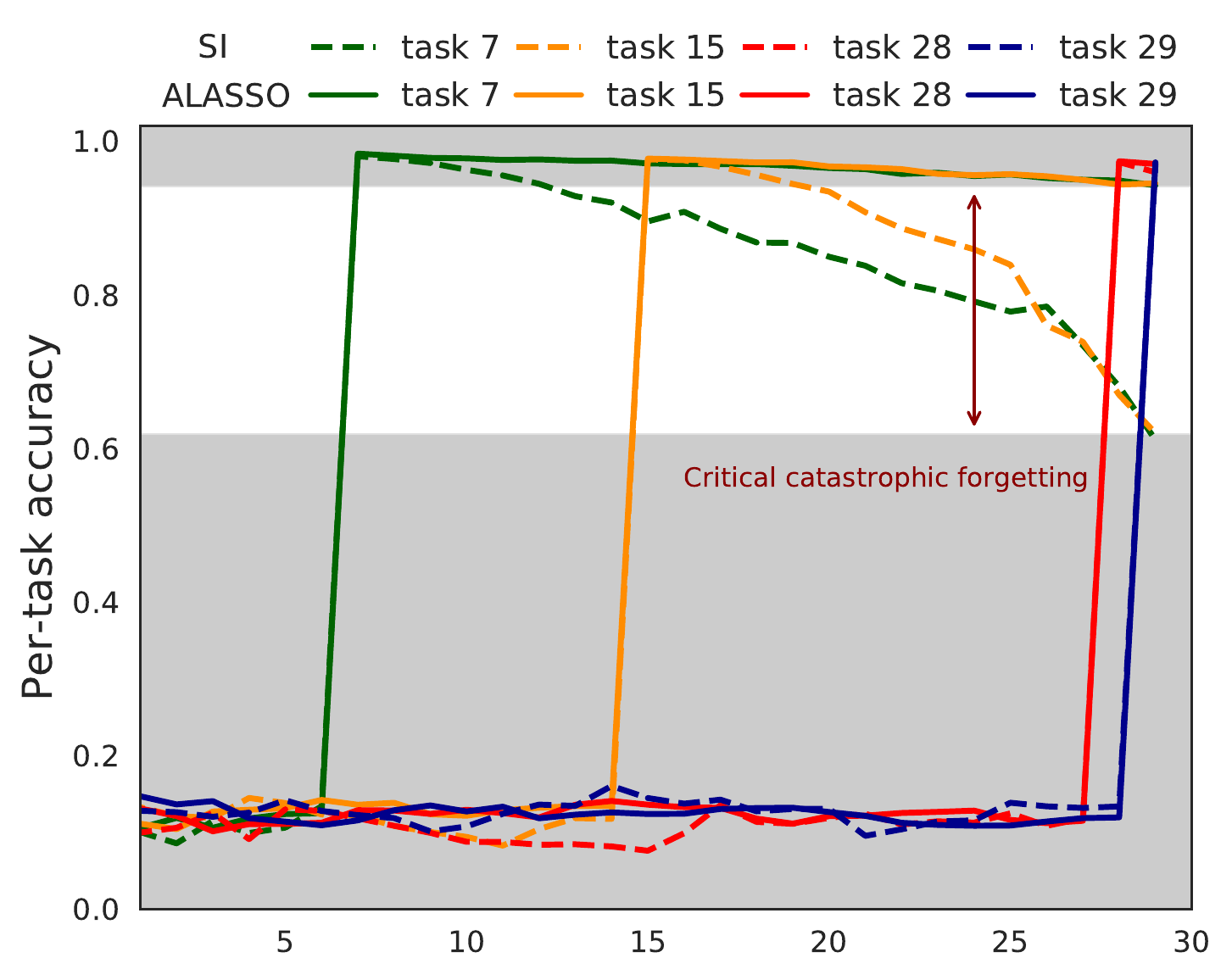}
	\end{center}
	\caption{Analysis of results from ALASSO in comparison to SI, which is one of the state-of-the-art methods. This figure presents per-task test accuracy of the selected tasks on the permuted MNIST dataset over time. In this graph, $x$-axis denotes the index of most recently trained task, and $y$-axis is accuracy.  
		All the visualized tasks achieve very high accuracy by the two algorithms when they are trained initially, but their accuracies are degraded as newer tasks come in.
		This forgetting problem is more catastrophic in SI than ALASSO by a large margin;
		SI is not effective to maintain accuracy of tasks learned in the past as the newer tasks are considered for training.}
	\vspace{-0.2cm}
	\label{fig:PreviousMethodProblemAnalysis}
\end{figure}

\subsection{Parameter Decoupling}
\label{sub:parameter}

The overestimated loss approximation is effective to reduce errors that happen inevitably in the previous tasks, but the optimization with this technique may suffer from slow convergence due to its inherent limitation.
Specifically, some hyperparameters, such as $a$ and $c$, affect the objective function in one way or another depending on where they occur in our formulation.
This is because we approximate the real loss for all the previous tasks using a single asymmetric function and it is almost impossible to consider all the possible combinations of the approximate functions estimated in the past.
\begin{figure*}[t]
	\begin{center}
		\includegraphics[width=0.49\textwidth]{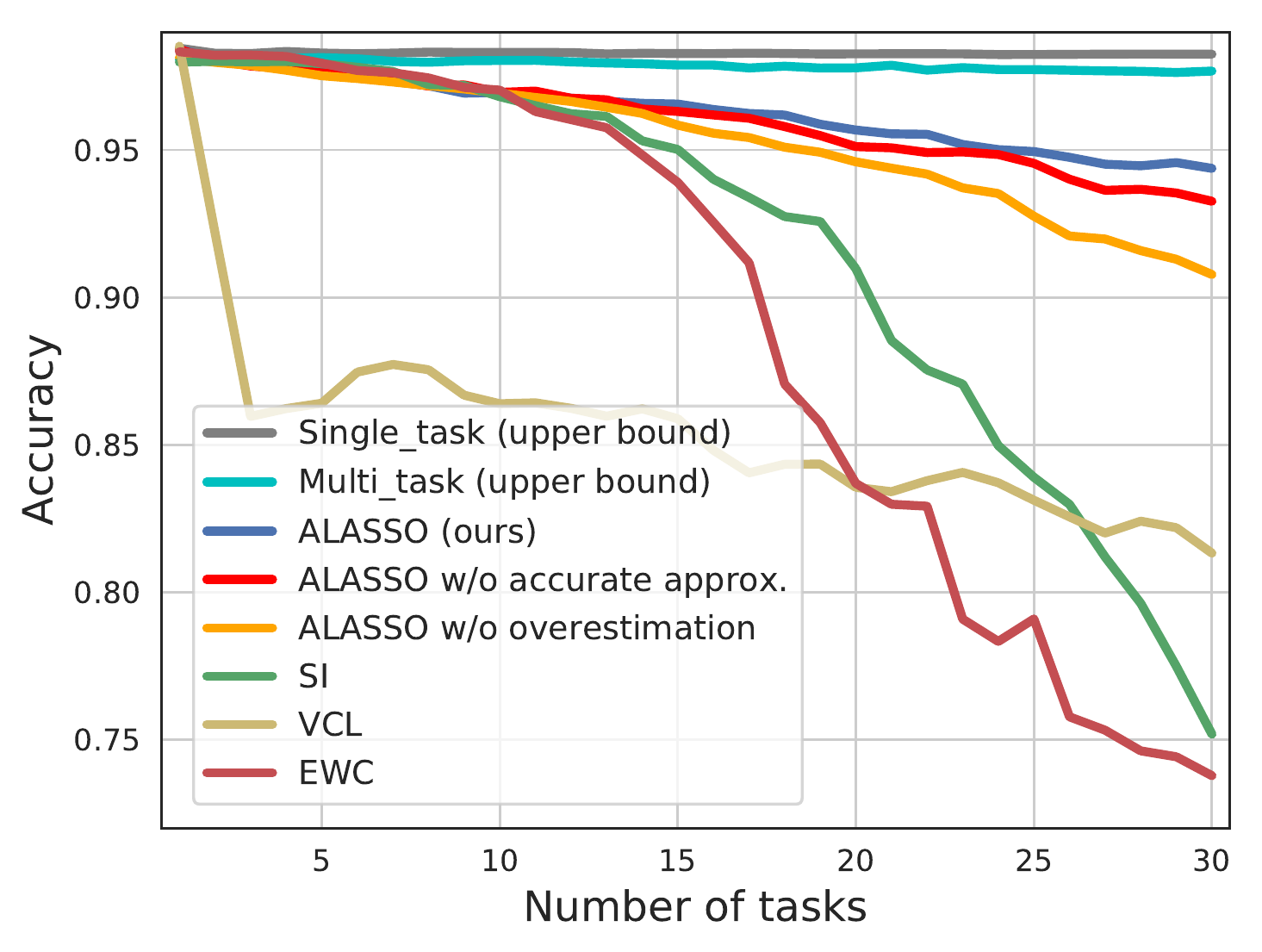}
		\hspace{0cm}
		\includegraphics[width=0.49\textwidth]{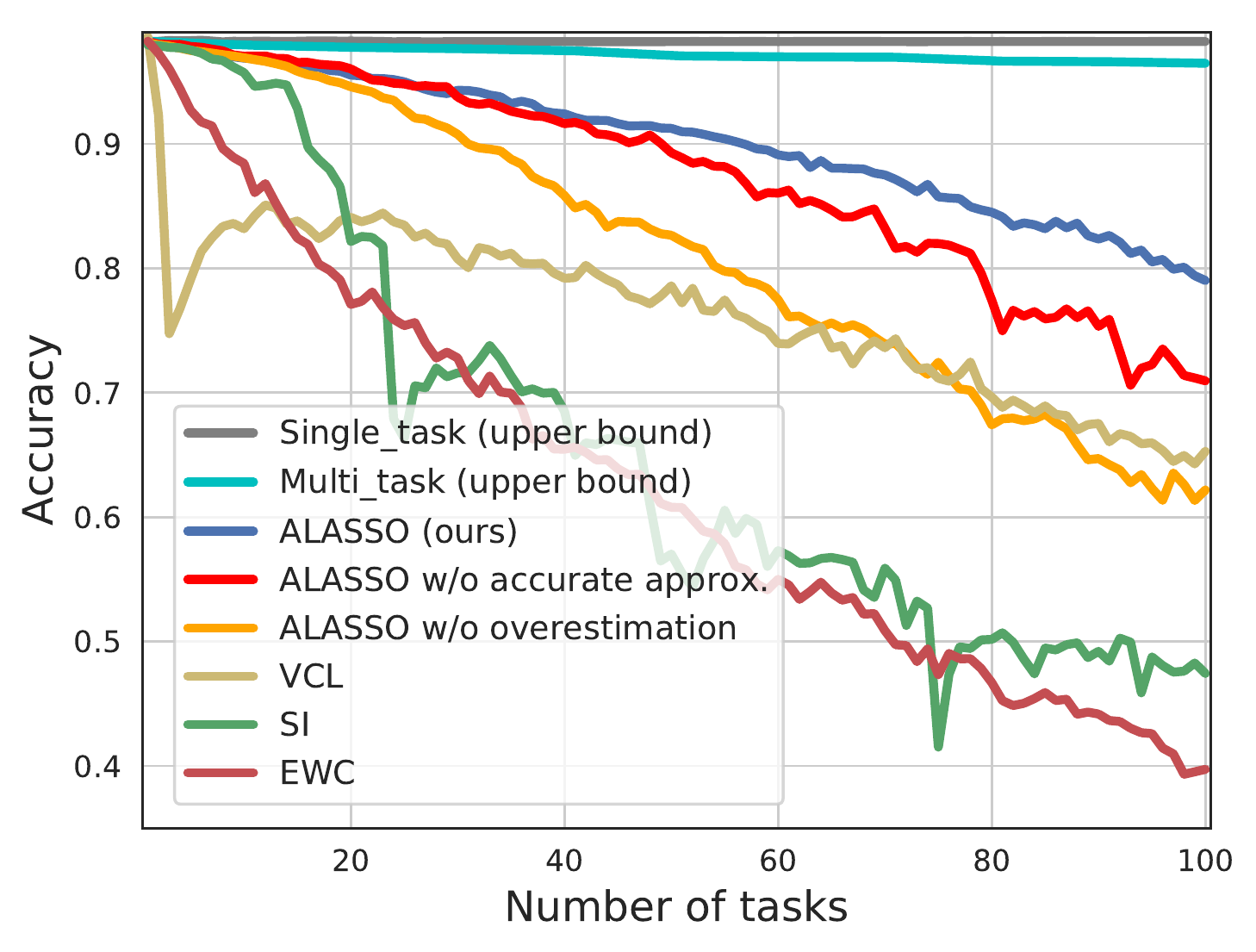}
	\end{center}
	\vspace{-0.2cm}
	\caption{Test accuracy on the permuted MNIST dataset with 30 (left) and 100 (right) tasks.  The results from several continual learning techniques including ours (ALASSO) are presented together with two upper-bound methods, {\it single\_task} and {\it multi\_task}. Note that $x$-axis denotes the index of a task given by the order of training. ALASSO achieves the state-of-the-art performance for both cases.  Each method uses the hyper-parameters optimized for itself. The accuracies of SGD, SGD+dropout, IMM and MAS are much worse and we do not include them in this graph for better visualization.}
	\vspace{-0.2cm}
	\label{fig:PermutedMNISTResult}
\end{figure*}

In our formulation, there are two different occurrences of hyperparameters; one is related to the gradient computation in SGD and the other is in calculating $\hat{\Omega}^n_k$ for loss approximation.
For example, increasing $c$ in Eq.~\eqref{eq:oala_surrogate} makes the model consider the previous tasks more while increasing $c$ in Eq.~\eqref{eq:past_omega} results in more weight on the current task by underestimating $\hat{\Omega}^n_k$.

To handle the inconsistent impact of identical hyperparameters on the optimization process, we decouple the parameters into two sets; a set of parameters used for computation of $\omega_k^{1:(n-1)}$ and the other set of parameters used for surrogate loss estimation.

Then, when we compute $\omega^{1:(n-1)}_k$ in Eq.~\eqref{eq:past_omega}, the hyperparameters $a$ and $c$ are decoupled from the other equations as
\begin{equation}
\omega^{1:(n-1)}_k = - c'\mathcal{L}_s^{n-1}\left(\hat{\theta}_k^{n},a'\right),
\end{equation}
where $a$ and $c$ in Eq.~\eqref{eq:past_omega} are replaced by $a'$ and $c'$, respectively.

\subsection{Discussion}
\label{sub:discussion}
Figure~\ref{fig:PreviousMethodProblemAnalysis} illustrates the promising results of the proposed algorithm in comparison to SI, which is one of the state-of-the-art methods. 
The figure presents how the accuracy of each task that is learned earlier changes over time as new tasks are added one by one.
We notice that the amount of degradation in ALASSO (ours) is much smaller than SI and the accuracy differences of the two algorithms at the same number of tasks are getting larger as time goes by.
These results clearly show the potential of our algorithm.

We claim that  the proposed algorithm is practically good because it does not involve side effects such as architectural modification, network size increase, additional memory requirements, batch processing, multiple execution of networks, and inference on multi-head networks.
We only need to store several variables such as $\Omega^n_k$, $\hat{\theta}^n_k$, and $\omega^n_k$ during training, and perform additional operations to compute the surrogate losses.

\section{Experiments}
\label{sec:experiment}

This section demonstrates performance of our algorithm compared to existing approaches on the standard datasets. 

\subsection{Datasets and Algorithms}
\label{sec:DatasetsandAlgorithms}
We employ three standard benchmark datasets to evaluate the proposed continual learning framework, which include the permuted MNIST, the split CIFAR-10/CIFAR-100 and the split Tiny ImageNet.
The permuted MNIST is a synthetic dataset based on MNIST~\cite{MNIST}, where all pixels of an image in MNIST are permuted differently but coherently in each task.
This dataset contains a large number of tasks and is widely used for evaluation of continual learning algorithms~\cite{CompeteToCompute, Goodfellow2013, EWC, OnlineEWC}.
The split CIFAR-10/CIFAR-100 dataset is generated from CIAFR-10 and CIFAR-100~\cite{CIFAR10and100} while the split Tiny ImageNet is derived from Tiny ImageNet~\cite{TinyImageNet}.
These datasets divide their target classes into multiple subsets, which correspond to individual tasks.

We compare our algorithm with the existing state-of-the-art methods including na\"ive SGD~\cite{robbins1951stochastic, kiefer1952stochastic}, SGD with dropout (SGD+dropout)~\cite{Goodfellow2013}, VCL~\cite{VCL}, EWC~\cite{EWC}, SI~\cite{OnlineEWC}, IMM~\cite{IMM} and MAS~\cite{MAS}.
To estimate the upper-bound accuracy of our approach, we present the results from the models learned for individual tasks and all the tasks in a batch, which are denoted by {\it single\_task} and {\it multi\_task}, respectively.

\begin{figure*}[t]
	\begin{minipage}{0.66\textwidth}
		\centering
		\includegraphics[width=0.485\textwidth]{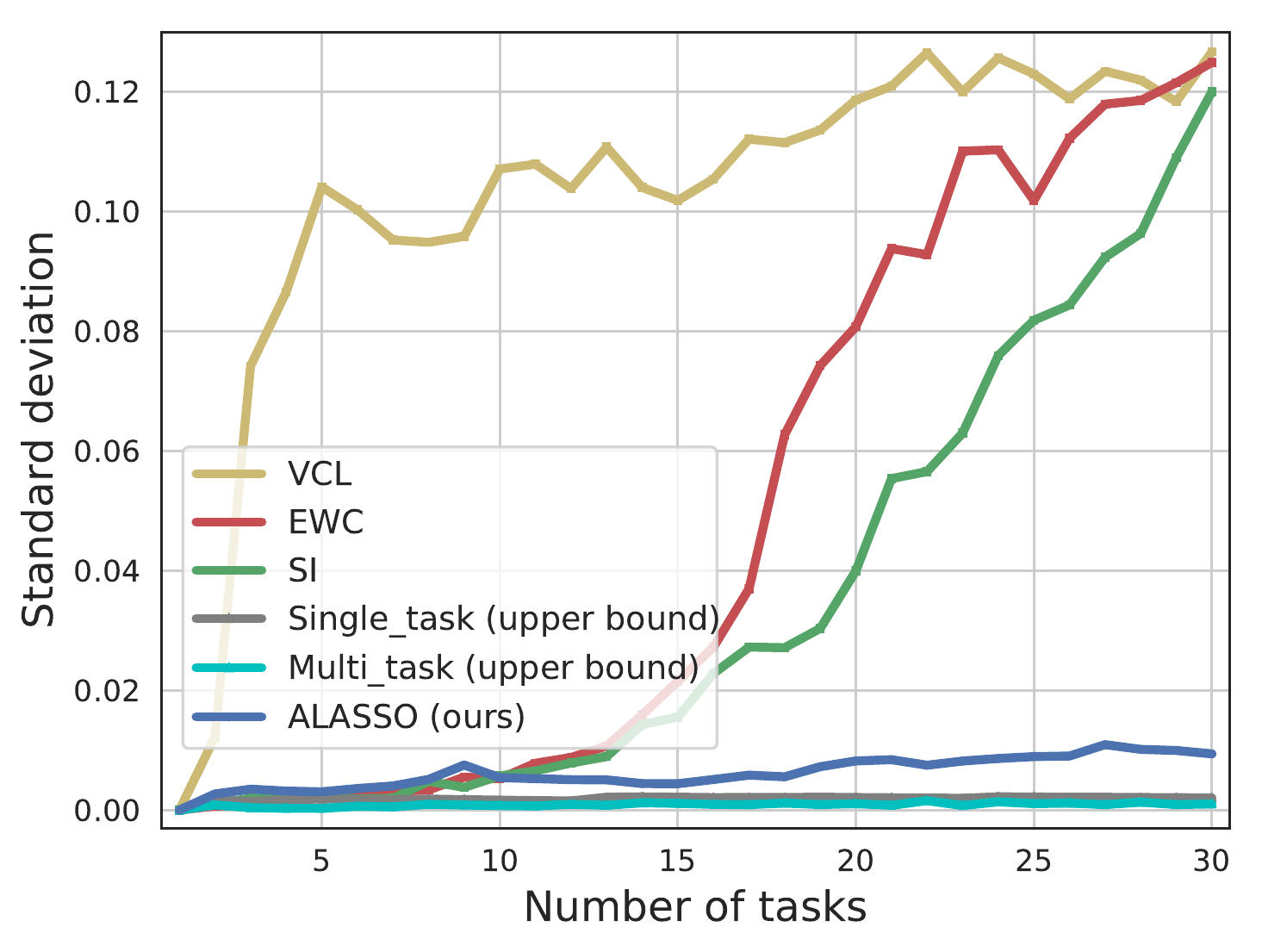}
		\hspace{0.1cm}
		\includegraphics[width=0.485\textwidth]{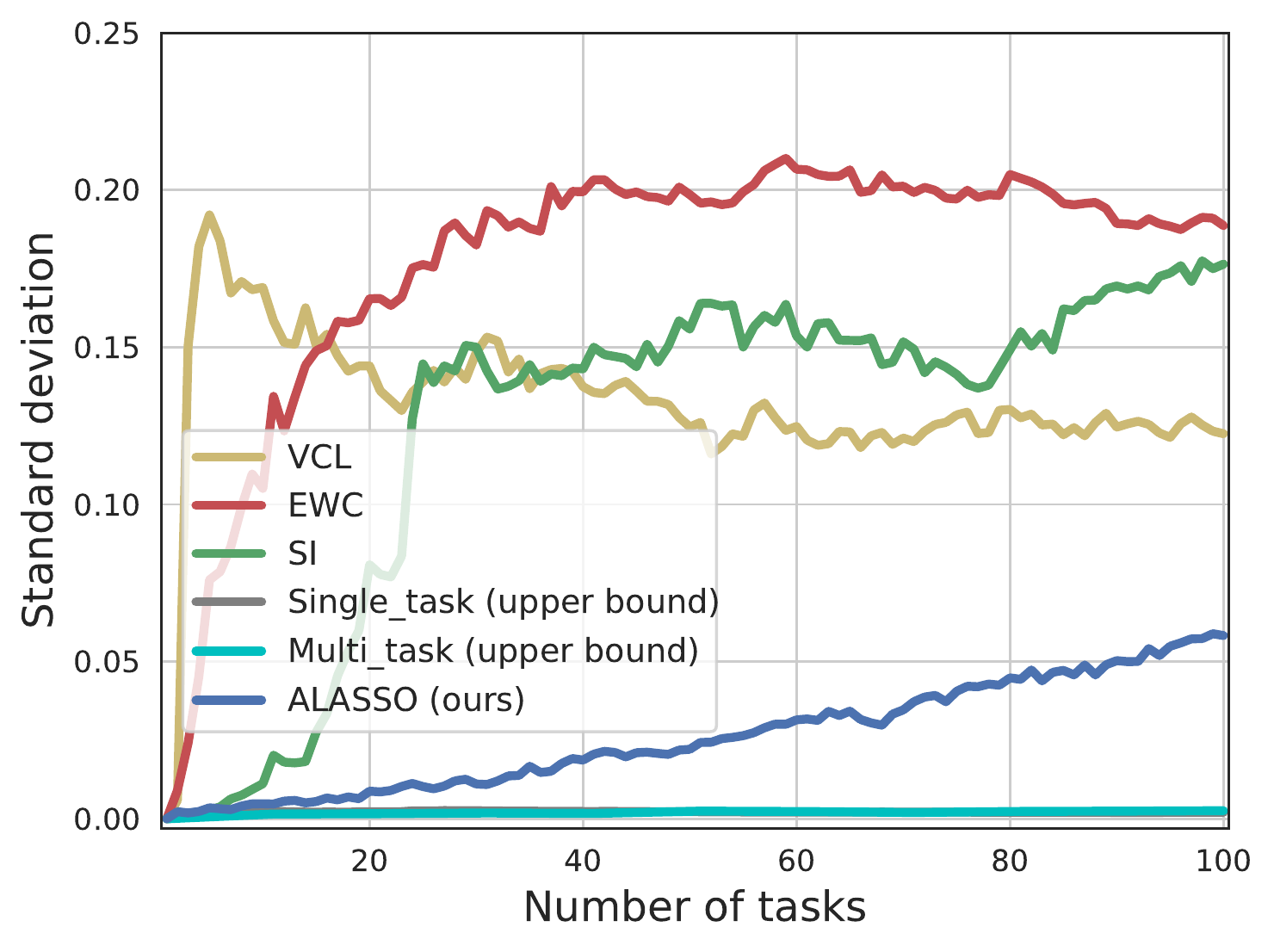}
		\caption{The standard deviation of per-task test accuracy on the permuted MNIST dataset when 30 (left) and 100 (right) tasks are learned online.}
		\label{fig:STD_PermutedMNIST}
	\end{minipage}\hfill
	\begin{minipage}{0.33\textwidth}
		\centering
		\includegraphics[width=0.97\textwidth]{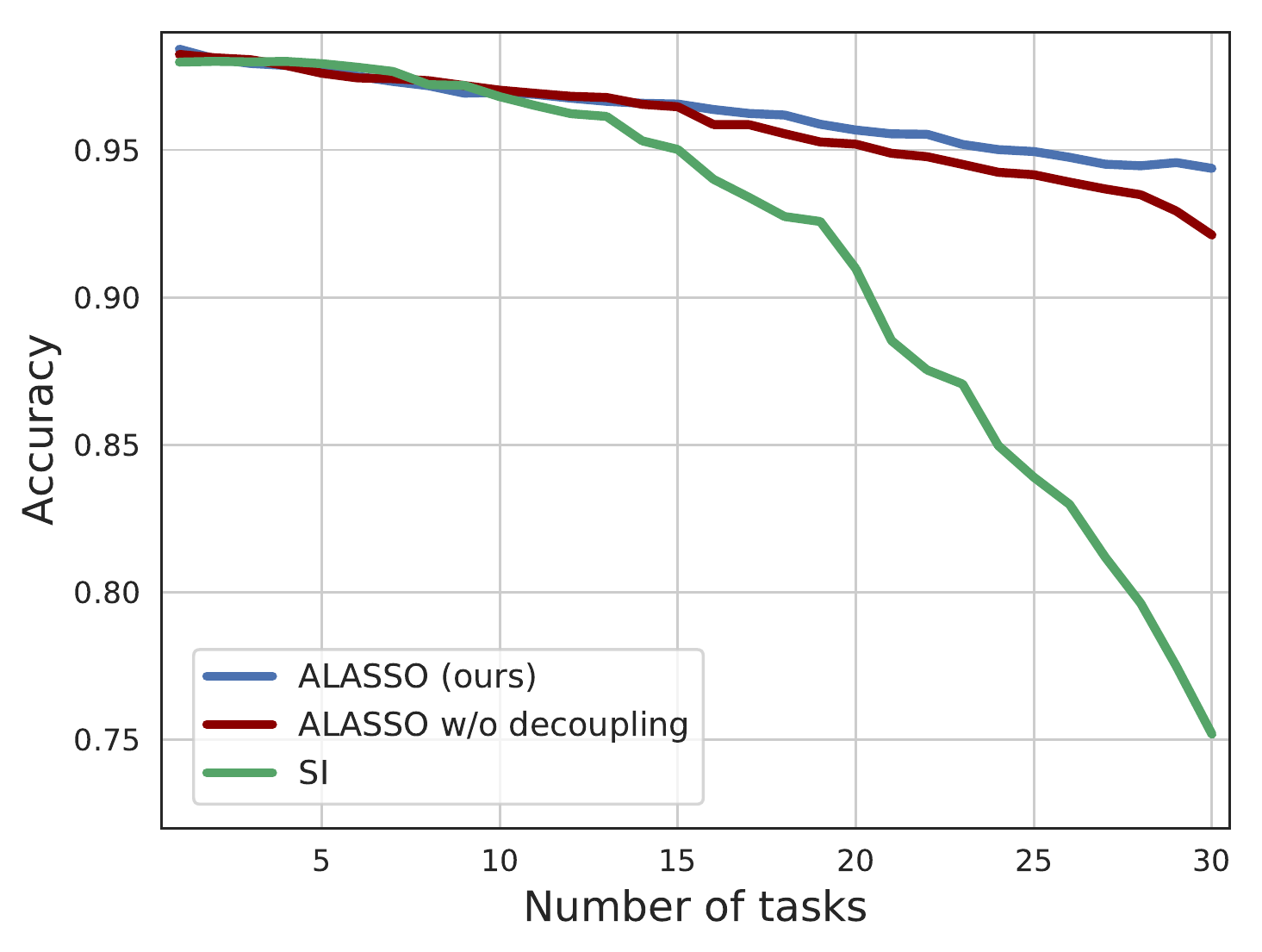}
		\caption{Result with and without parameter decoupling on the permuted MNIST.}
		\label{fig:DecouplingResult}
	\end{minipage}
\end{figure*}

\subsection{Training Details}
\label{sub:training}
Our models are trained based on the description in SI~\cite{OnlineEWC}; the models are optimized by Adam and the learning rate is 0.001 for all the tested datasets.
The batch size is 256 for the permuted MNIST and the split CIFAR-10/CIFAR-100, and 128 for the split Tiny ImageNet while the numbers of epochs are 20 for the permuted MNIST and 60 for the other two datasets.
We plan to release our source code and raw results for better reproducibility.

\subsection{Results on the Permuted MNIST}
\label{sec:PermutedMNIST}

We present the results from all compared algorithms on 30 and 100 tasks of the permuted MNIST dataset\footnote{For better visualization, we omit results from SGD, SGD+dropout, IMM and MAS because their accuracies are significantly lower than the presented ones.}.
The network architectures used in this experiment are simple multi-layer perceptrons composed of two hidden layers with additional ReLU and softmax output layers.
Dropout is not used except for SGD+dropout.

As Figure~\ref{fig:PermutedMNISTResult} shows, ALASSO outperforms all compared methods with large margins (about 15\% point at least) for both 30 and 100 tasks.
Note that all other methods undergo large performance drops as the number of tasks increases while ALASSO is more robust than others; it presents only moderate performance loss even after learning 100 tasks.

To verify the effectiveness of the asymmetric loss overestimation on the unobserved side and the accurate quadratic approximation on the observable one, we perform the ablation study to analyze their benefit.
Figure~\ref{fig:PermutedMNISTResult} also illustrates the clear contribution of the components.
%

Figure~\ref{fig:STD_PermutedMNIST} shows that the accuracy of ALASSO is stable compared to the other methods.
The overall performance of ALASSO is not affected by parameter decoupling but rather improves accuracy slightly as illustrated in Figure~\ref{fig:DecouplingResult}.
Note that it facilitates the fast convergence of the models.

\subsection{Results from the Split CIFAR-10/CIFAR-100}

\begin{figure}[t]
	\vspace{-0.1cm}
	\begin{center}
		\includegraphics[width=\linewidth]{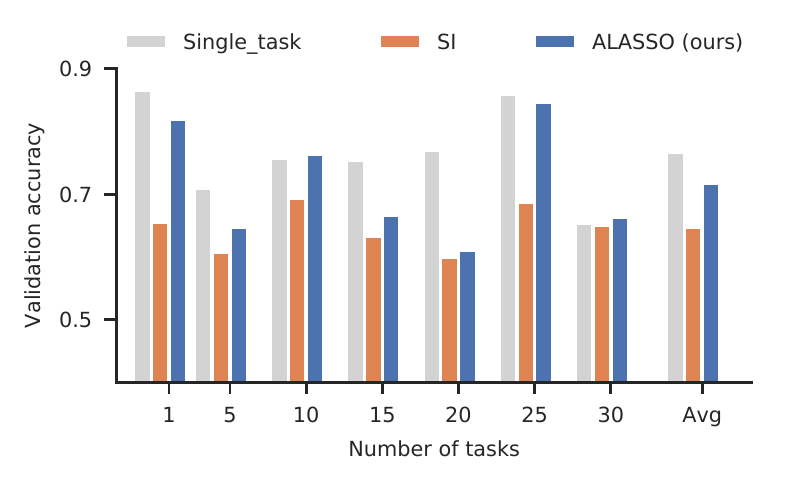}
	\end{center}
	\vspace{-0.3cm}
	\caption{Per-task accuracy at every 5 task on the split CIFAR-10/CIFAR-100 when training 30 tasks in an online manner. ALASSO works better than SI and comparable to {\it single\_task} in this more realistic dataset.}
	\vspace{-0.2cm}
	\label{fig:CifarResult}
\end{figure}

We evaluate our algorithm on a more realistic scenario using the split CIFAR-10/CIFAR-100 dataset.
In this experiment, each task is composed of 3 classes and the number of classes in a task gets larger as the index of the task increases. 
Our model is based on a convolutional neural network with 4 convolutional layers followed by 2 fully connected layers with dropouts; ReLU and $2 \times 2$ max pooling are also employed to add nonlinearity in the model.
%
Figure~\ref{fig:CifarResult} demonstrates the results from {\it single\_task} (to show the upper-bound accuracy), SI, and ALASSO on the split CIFAR-10/CIFAR-100 dataset.
The proposed method outperforms SI consistently even on this more realistic dataset and achieves 5.7\% point better than SI in average; the accuracy of SI drops by 13.9\% point and 41\% of the lost accuracy in SI is recovered by ALASSO.
Note that ALASSO even shows the comparable performance to {\it single\_task} method in some cases.

\begin{figure}[t]
	\vspace{-0.1cm}
	\begin{center}
		\includegraphics[width=\linewidth]{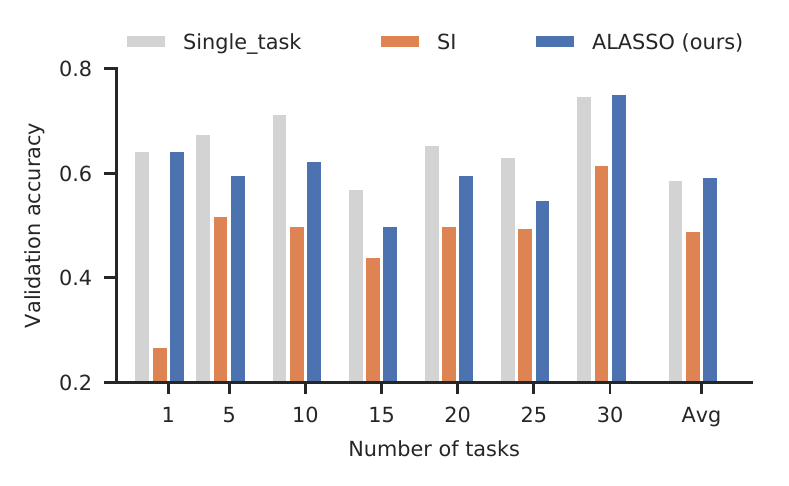}
	\end{center}
	\vspace{-0.3cm}
	\caption{Per-task accuracy of every 5 task on the split Tiny ImageNet when training 30 tasks in an online manner.  Even in this more realistic dataset with more classes, the accuracy ALASSO is better than SI and even comparable to {\it single\_task}.}
	\label{fig:TinyImageNetResult}
\end{figure}

\subsection{Results from the Split Tiny ImageNet}

We conduct an experiment with a larger network on the split Tiny ImageNet dataset, which has more classes and consists of the tasks with 6 classes.
The model has 5 convolutional layers followed by 2 fully connected layers with dropouts while $2 \times 2$ max pooling.
ReLU activation function are employed as well to add nonlinearity of the network.

Figure~\ref{fig:TinyImageNetResult} illustrates the results from {\it single\_task}, SI, and ALASSO.
Note that the overall accuracy of ALASSO is as competitive to the \textit{single\_task}, which is supposed to show a practical upper-bound performance conceptually.
In reality, the average validation accuracy of our method is 59.4\% and even higher than \textit{single\_task} accuracy, 58.9\%.
ALASSO also outperforms SI by approximately $10\%$ point in average.

\subsection{More Analysis}
\begin{table}[t]
	\centering
	\caption{Sensitivity to $a$ on the permuted MNIST}
	\vspace{0.1cm}
\scalebox{0.85}{
		\begin{tabular}{c | c c | c c c c} 
			$a$ &0.8 & 1.0  & 2.0 &  3.0  & 4.0 & 5.0 \\  
			\hline\hline
			30 tasks &0.91 & 0.92 &  0.94   & 0.94 & 0.94  & 0.94 \\
			100 tasks & 0.59&0.62 & 0.79  & 0.79  & 0.78  & 0.75 \\
			\hline
		\end{tabular}
}	
	\vspace{-0.2cm}
	\label{table:sensitivity}
\end{table}
The primary hyperparameter in our algorithm is $a$, which is introduced to overestimate the unobserved side of the loss function. Since $a$ determines the factor of overestimation, it is reasonable to set its value larger than 1.0.
We determine the value of $a$ empirically in the permuted MNIST dataset; we choose the similar values for the other datasets, and they are fixed within each dataset. 
According to our experience, the overall performance is not sensitive to a wide range of $a$'s value as presented in Table~\ref{table:sensitivity}. 
if $a$ is 1, loss function is approximated by symmetric function.
Note that the performance of our algorithm degrades when $a=0.8~(<1)$.
Another hyperparameter $c$ balances between the losses in the current and the past task, and it is set to 1.0 throughout the experiments.

%
\begin{figure}[t]
	\vspace{-0.1cm}
	\begin{center}
		\includegraphics[width=\linewidth]{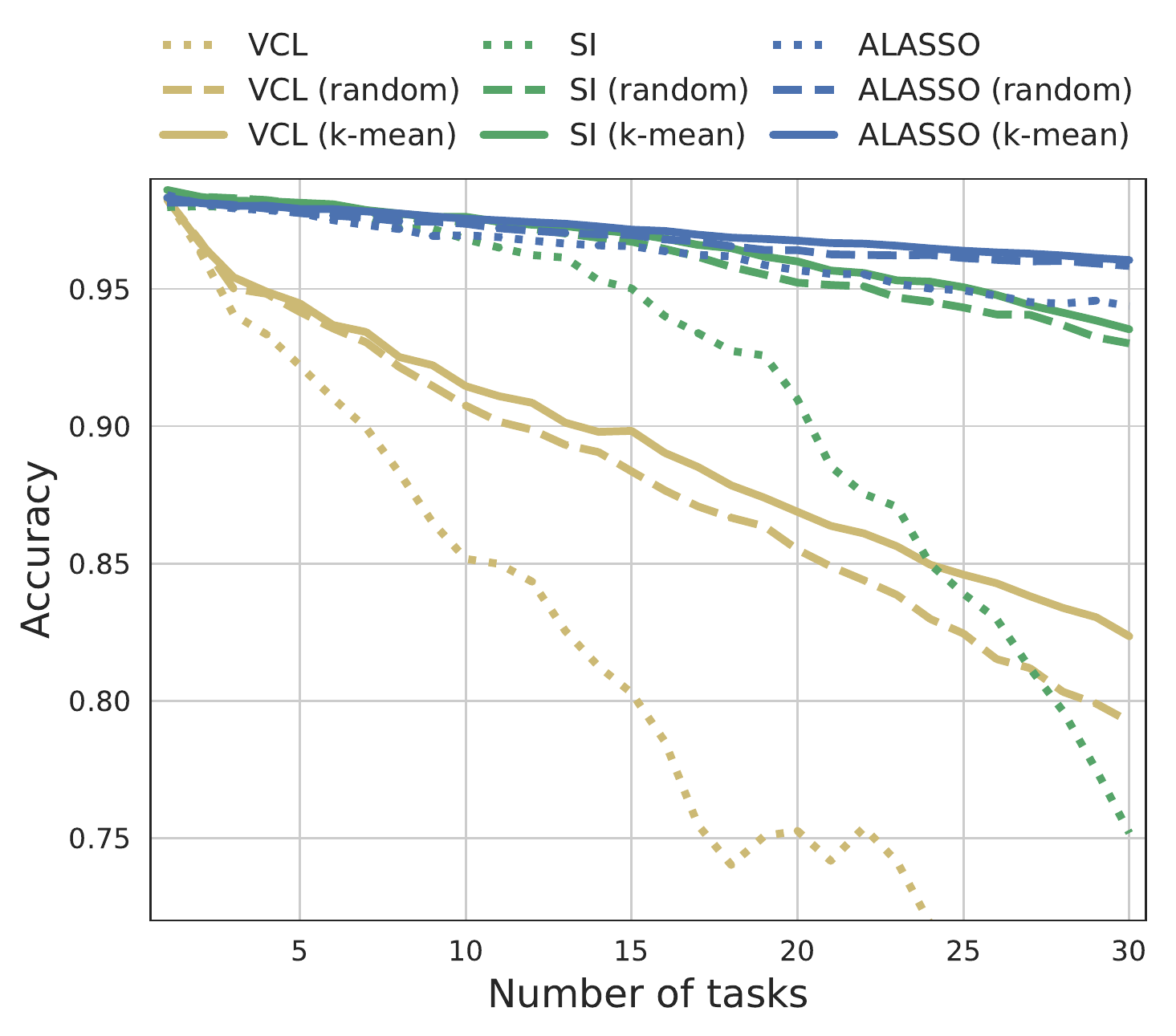}
	\end{center}
	\vspace{-0.4cm}
	\caption{Results with episodic memory for 200 examples per task on the permuted MNIST.}
	\label{fig:memory}
\end{figure}

Although our original formulation uses no additional memory, the performance of ALASSO is compared with VCL~\cite{VCL}, which is a method with episodic memory, in the same condition.
Figure~\ref{fig:memory} illustrates that our algorithm still achieve outstanding performance.

\begin{table}[t]
	\centering 
	\caption{Algorithm comparison in terms of accuracy, forgetting and intransigence measures~\cite{reimannian}}
	\vspace{0.1cm}
	\scalebox{0.85}{
		\begin{tabular}{c | c c c | c c c} 
			&\multicolumn{3}{c|}{Permuted MNIST}  & \multicolumn{3}{c}{Split CIFAR10/100}   \\
			Measure& $A_{30}$ & $F_{30}$ & $I_{30}$ & $A_{30}$ & $F_{30}$ & $I_{30}$ \\
			\hline\hline
			EWC & 0.738 & 0.243 & 0.010 &- & -& -\\
			SI & 0.752 & 0.232 & \textbf{0.008} & 0.632& 0.110 & 0.137\\
			Ours & \textbf{0.944} & \textbf{0.027} & 0.015& \textbf{0.697}& \textbf{0.026} & \textbf{0.123} \\ [0.5ex] \hline 
	\end{tabular}}
	\vspace{-0.2cm}
	\label{table:measure} 
\end{table}

Table~\ref{table:measure} present accuracy ($A$), forgetting ($F$) and intransigence ($I$) measures of EWC, SI, and ALASSO on the permuted MNIST and the split CIFAR-10/CIFAR-100, where the subscripts besides $A$, $F$, and $I$ denote the number of tasks when the performance is computed.
Note that the accuracy measure is most comprehensive and is conceptually correlated to the other two ones.
ALASSO outperforms EWC and SI, especially in terms of accuracy and forgetting measures.
\section{Conclusion}
\label{sec:conclusion}

We presented a novel continual learning framework based on the overestimated asymmetric loss approximation with the better parametrization for the quadratic approximation, which is a carefully designed generalized version of SI.
Our algorithm alleviates the catastrophic forgetting issue, which is common in deep neural networks, and is particularly helpful for the scenario with a large number of tasks in continual learning.
The proposed solution is motivated by the observation that network parameter updates do not affect target loss function symmetrically, and does not incur substantial side effects.
It achieves the state-of-the-art performance on several challenging standard benchmark datasets.

\vspace{-0.2cm}
\paragraph*{Acknowledgments}
This work was supported by the National Research Foundation (NRF) grant funded by the Korea Government (MSIT) (NRF2017R1A2B2011862).

{\small
	\bibliographystyle{ieee_fullname}
	\bibliography{egbib}
}


\makeatletter
\newcommand{\thickhline}{%
	\noalign {\ifnum 0=`}\fi \hrule height 1.3pt
	\futurelet \reserved@a \@xhline
}

\renewcommand{\thesection}{\Alph{section}}
\renewcommand{\thefigure}{\Alph{figure}}
\renewcommand{\thetable}{\Alph{table}}
\renewcommand{\theequation}{A.\arabic{equation}}


\onecolumn

\textbf{\textsc{\LARGE Supplementary Material}}\\

This document discusses intuitive interpretation of  $\mathcal{L}^n(\theta_k)$ in Section~\ref{sec:AdditionComment}.
Section~\ref{sec:Derivation} presents the derivation of $\hat{\Omega}_k^n$ which is quadratic approximation parameter in SI~\cite{OnlineEWC} and ALASSO(ours). Section~\ref{sec:AdditionComparison} presents the results on 30 tasks of permuted MNIST dataset from all state-of-the-art comparable algorithms. Section~\ref{sec:Configuration} provides the detailed configuration of the experimental network architectures.

\section{Additional comment of $\mathcal{L}^n(\theta_k)$}
\label{sec:AdditionComment}
This section presents loss for the current task $\mathcal{L}^n(\theta_k)$, which is a component dependent on the parameter $\theta_k$ in loss $\mathcal{L}^n(\theta)$.
The loss $\mathcal{L}^n(\theta)$ can be expressed by the sum of $\mathcal{L}^n(\theta_k)$:

\begin{align}
\mathcal{L}^n(\theta)=\sum_k \mathcal{L}^n(\theta_k).
\label{eq:Sum}
\end{align}
$\mathcal{L}^n(\theta_k)$ can be interpreted as the parameter specific contribution to the total loss $\mathcal{L}^n(\theta)$.


\hfill

\section{Derivation of $\hat{\Omega}_k^n$ in Eq.~(3) and (7) in the main paper}
\label{sec:Derivation}
Assuming that the surrogate loss function for each parameter up to the $n^\text{th}$ task, $\mathcal{L}_s^n(\theta_k)$, is defined by a quadratic function, which is further decomposed of the two terms as
\begin{align}
\mathcal{L}_s^n(\theta_k) \equiv \hat{\Omega}^{n}_k\left(\theta_k - \hat{\theta}^{n}_k \right)^2
= \mathcal{L}^n\left(\theta_k\right) +c\mathcal{L}_s^{n-1}(\theta_k),
\label{eq:loss_sup}
\end{align}
where $\mathcal{L}^n\left(\theta_k\right)$ is the loss for the current task, $\mathcal{L}_s^{n-1}(\theta_k)$ is the surrogate loss function up to $\left(n-1\right)^\text{th}$ task, and $c$ is a hyperparameter to balance between the two terms.

After completing learning of the $n^\text{th}$ task, we can further assume that we already have $n^\text{th}$ loss function $\mathcal{L}^n\left(\theta_k\right)$ and the new model parameter $\hat{\theta}^{n}_k$ given $\hat{\Omega}^{n-1}_k$ and $\hat{\theta}^{n-1}_k$ from the previous iteration.
Then, we can derive the value of $\hat{\Omega}^{n}_k$ satisfying Eq.~\eqref{eq:loss_sup} in SI and ALASSO based on the following procedures.

\subsection{Synaptic Intelligence ~\cite{OnlineEWC}}
Eq.\eqref{eq:Omega_si} shows that the $\hat{\Omega}^n_k$ obtained by the SI ~\cite{OnlineEWC} method is not accurate approximation for quadratic surrogate loss function.
\begin{align} 
\hat{\Omega}^n_k &= \frac{\mathcal{L}^n\left(\theta_k\right)+c\mathcal{L}_s^{n-1}\left(\theta_k\right)}{\left(\theta_k - \hat{\theta}_k^n\right)^2} &\qquad & \text{from Eq.~\eqref{eq:loss_sup}} \nonumber \\
&= \frac{\mathcal{L}^n\left(\theta_k\right)+c\hat{\Omega}^{n-1}_k\left(\theta_k - \hat{\theta}_k^{n-1}\right)^2}{\left(\theta_k - \hat{\theta}_k^n\right)^2}  &\qquad & \text{by plugging in $\hat{\theta}_k^{n-1}$} \nonumber\\
&= \frac{\mathcal{L}^n\left(\hat{\theta}_k^{n-1}\right)-\mathcal{L}^n\left(\hat{\theta}_k^n\right)}{\left(\hat{\theta}_k^{n-1} - \hat{\theta}_k^n\right)^2}+\frac{\mathcal{L}^n\left(\hat{\theta}_k^n\right)}{\left(\hat{\theta}_k^{n-1} - \hat{\theta}_k^n\right)^2} & \nonumber \\
&= \frac{\mathcal{L}^n\left(\hat{\theta}_k^{n-1}\right)-\mathcal{L}^n\left(\hat{\theta}_k^n\right)}{\left(\hat{\theta}_k^{n-1} - \hat{\theta}_k^n\right)^2}+\frac{\mathcal{L}^n\left(\hat{\theta}_k^n\right)\cdot\hat{\Omega}^{n-1}_k }{\left(\hat{\theta}_k^{n-1} - \hat{\theta}_k^n\right)^2\cdot\hat{\Omega}^{n-1}_k} & \nonumber \\
&= \frac{\mathcal{L}^n\left(\hat{\theta}_k^{n-1}\right)-\mathcal{L}^n\left(\hat{\theta}_k^n\right)}{\left(\hat{\theta}_k^{n-1} - \hat{\theta}_k^n\right)^2}+\frac{\mathcal{L}^n\left(\hat{\theta}_k^n\right)\cdot\hat{\Omega}^{n-1}_k }{\mathcal{L}_s^{n-1}\left(\hat{\theta}_k^n\right)} & \nonumber \\
&=\frac{\omega^n_k }{\left(\hat{\theta}_k^{n-1} - \hat{\theta}_k^n\right)^2}+\frac{\mathcal{L}^n\left(\hat{\theta}_k^n\right)\cdot\hat{\Omega}^{n-1}_k }{\mathcal{L}_s^{n-1}\left(\hat{\theta}_k^n\right)} &\qquad & \text{by introducing a new variable $\omega_k^n$} \nonumber \\
&\approx \frac{\omega^n_k }{\left(\hat{\theta}_k^{n-1} - \hat{\theta}_k^n\right)^2}+\hat{\Omega}^{n-1}_k, &
\label{eq:Omega_si}
\end{align}
where $\omega^n_k$ denotes the difference between the losses for the task $n$ before and after training the task , $\omega^n_k=\mathcal{L}^n\left(\hat{\theta}_k^{n-1}\right)-\mathcal{L}^n\left(\hat{\theta}_k^n\right)$


\newpage
\subsection{ALASSO}
Eq.~\eqref{eq:Omega_alasso} shows that the $\hat{\Omega}^n_k$ obtained by ALASSO provides the perfect quadratic approximation.
\begin{align} 
\hat{\Omega}^n_k &= \frac{\mathcal{L}_s^n\left(\theta_k\right)}{\left(\theta_k - \hat{\theta}_k^n\right)^2} &\qquad & \text{from Eq.~\eqref{eq:loss_sup}}  \nonumber \\ 
&= \frac{\mathcal{L}_s^n \left(\hat{\theta}_k^{n-1}\right)}{\left(\hat{\theta}_k^{n-1} - \hat{\theta}_k^n \right)^2}  &\qquad & \text{by plugging in $\hat{\theta}_k^{n-1}$}\nonumber \\
&= \frac{\mathcal{L}_s^{n}\left(\hat{\theta}_k^{n-1}\right)- \hat{\Omega}^n_k\left(\hat{\theta}_k^n - \hat{\theta}_k^n \right)^2}{\left(\hat{\theta}_k^{n-1} - \hat{\theta}_k^n \right)^2} \nonumber \\
&= \frac{\mathcal{L}_s^{n}\left(\hat{\theta}_k^{n-1}\right)- \mathcal{L}_s^{n}\left(\hat{\theta}_k^{n}\right)}{\left(\hat{\theta}_k^{n-1} - \hat{\theta}_k^n \right)^2} \nonumber \\
&= \frac{\left(\mathcal{L}^{n}\left(\hat{\theta}_k^{n-1}\right)+c\mathcal{L}_s^{n-1}\left(\hat{\theta}_k^{n-1}\right) \right)-\left(\mathcal{L}^{n}\left(\hat{\theta}_k^{n}\right)+c\mathcal{L}_s^{n-1}\left(\hat{\theta}_k^{n}\right) \right)}{\left(\hat{\theta}_k^{n-1} - \hat{\theta}_k^n \right)^2} &\qquad & \text{from Eq.~\eqref{eq:loss_sup}}  \nonumber  \\
&= \frac{\left(\mathcal{L}^{n}\left(\hat{\theta}_k^{n-1}\right)-\mathcal{L}^{n}\left(\hat{\theta}_k^{n}\right) \right)+c\left(\mathcal{L}_s^{n-1}\left(\hat{\theta}_k^{n-1}\right)-\mathcal{L}_s^{n-1}\left(\hat{\theta}_k^{n}\right) \right)}{\left(\hat{\theta}_k^{n-1} - \hat{\theta}_k^n \right)^2}  \nonumber \\
&= \frac{\omega^n_k + \omega^{1:(n-1)}_k}{(\hat{\theta}_k^n -\hat{\theta}_k^{n-1})^2}. &\qquad & \text{by introducing a new variable $\omega_k^n$, $\omega_k^{1:(n-1)}$}
\label{eq:Omega_alasso}
\end{align}
where $\omega^n_k$ denotes the difference between the losses for the task $n$ before and after training the task, $\omega^n_k=\mathcal{L}^n\left(\hat{\theta}_k^{n-1}\right)-\mathcal{L}^n\left(\hat{\theta}_k^n\right).\ $
$\omega_k^{1:(n-1)}$ denotes the difference between the surrogate losses of previous tasks for the task $n$ before and after training the task, $\omega_k^{1:(n-1)}=c\left(\mathcal{L}^{n-1}_s\left(\hat{\theta}_k^{n-1}\right)-\mathcal{L}_s^{n-1}\left(\hat{\theta}_k^n\right)\right)$.

\newpage
\section{Additional comparison with other methods}
\label{sec:AdditionComparison}
\begin{figure*}[h]
\begin{center}
\includegraphics[width=0.65\linewidth]{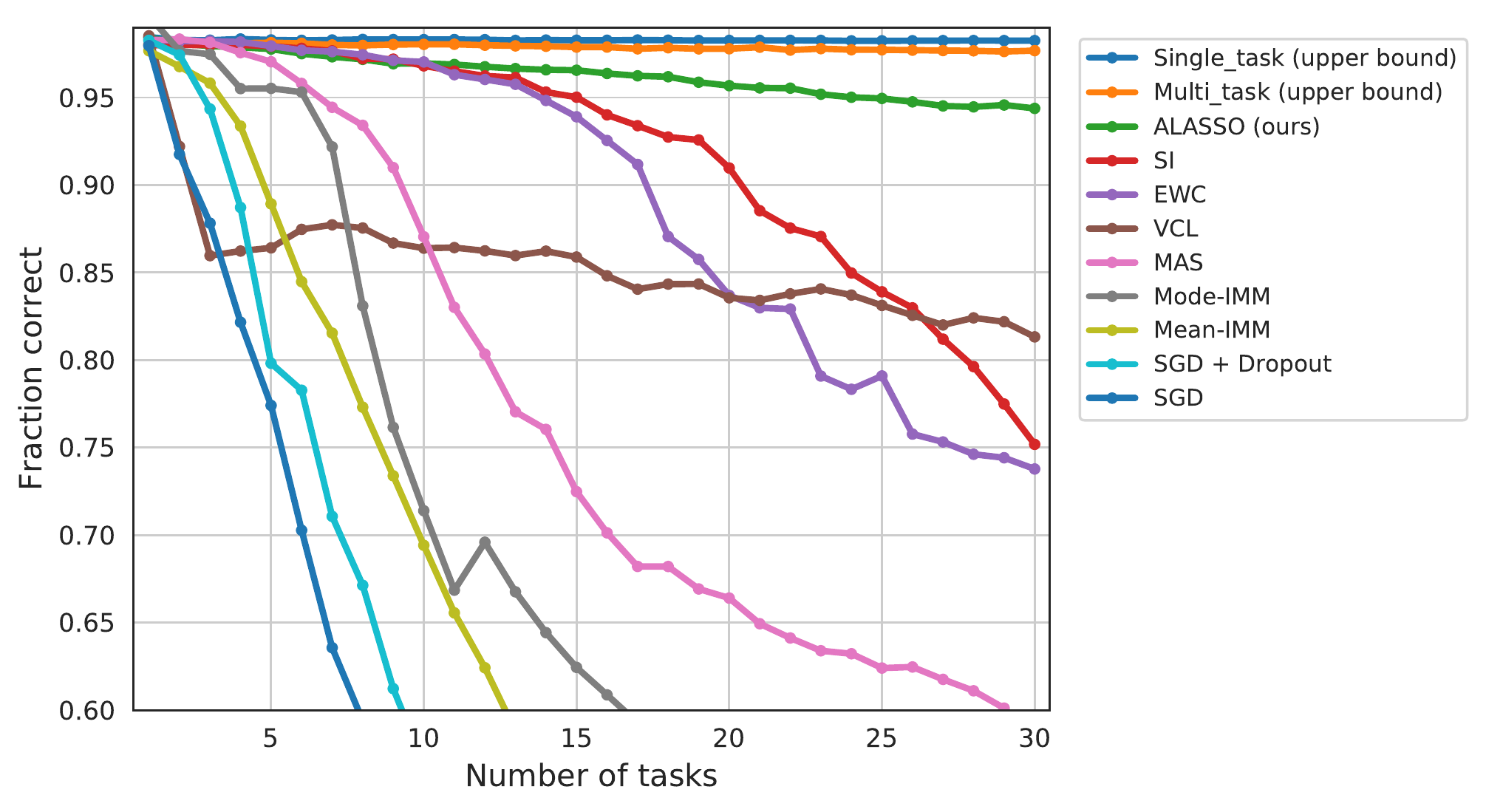}
\end{center}
\caption{Total test accuracy of all previously and currently learned tasks on permuted MNIST (one of the most commonly used benchmarks) over time with different continual learning methods (including our method, ALASSO), and single task and multi task (as upper bounds) (x-axis: the index of the lastly trained task. Our method achieves state-of-the-art performance for 30 tasks (even with a near-upper-bound result)}
\label{fig:SupplePMNIST}
\end{figure*}

\begin{figure*}[h]
\begin{center}
\includegraphics[width=0.65\linewidth]{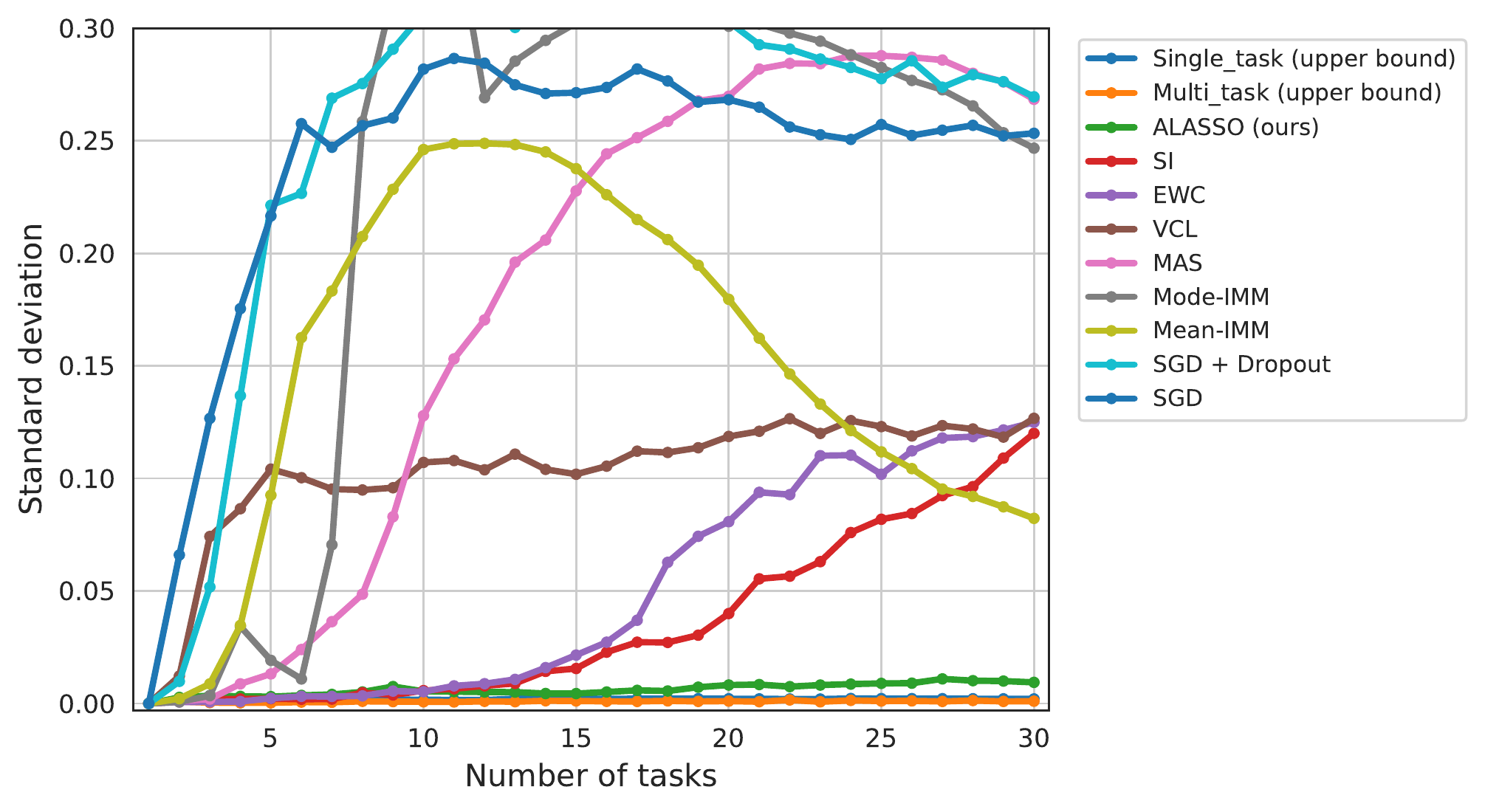}
\end{center}
\caption{Standard deviation of per-task test accuracy of all previously and currently learned tasks on permuted MNIST over time with different continual learning methods (including our method, ALASSO) and single task and multi task (as upper bounds) (x-axis: the index of the lastly trained task. Our method achieves near-0 standard deviation similar to the upper bound)}
\label{fig:SuplePMNISTSTD}
\end{figure*}

In this subsection we show the comparative experimental results with more algotithms on the permuted MNIST 30 tasks. 
The additionally compared with the existing state-of-the-art algorithms include SI~\cite{OnlineEWC}, EWC~\cite{EWC}, VCL~\cite{VCL}, MAS~\cite{MAS}, IMM~\cite{IMM}, na\"ive stochastic gradient descent (SGD)~\cite{robbins1951stochastic,kiefer1952stochastic}, SGD with dropout (SGD+dropout) ~\cite{Goodfellow2013}. 
Figure~\ref{fig:SupplePMNIST} and Figure~\ref{fig:SuplePMNISTSTD} show the accuracy and standard deviation of each algorithm over the number of tasks.
The \emph{Multi task} 
as another upper bound means a simple multi-task learning with a mini-batch mixing data of all the previous tasks and the current task. According to experiment in Figure \ref{fig:SupplePMNIST}, our algorithm achieves the best performance among all compared methods with large margins (about 15\% point at least). Besides, Figure \ref{fig:SuplePMNISTSTD} shows that our algorithm is stable in the sense that the performance variation across tasks are small, compared to all other methods, even though the learning process is incremental.

\newpage
\section{Configuration of the experimental architectures}
\label{sec:Configuration}
The detail network architecture configuration information for our experiments on the Permuted MNIST ( Table. \ref{table:PermutedMNIST}) / Split CIFAR-10,CIFAR-100 (Table. \ref{table:CIFAR}) /Tiny ImageNet (Table. \ref{table:TinyImageNet}) are given below.

\begin{table*}[h]
	\centering
	\caption{Network Architecture for Permuted MNIST Experiment} \vspace{0.2cm}
	\scalebox{0.95}{
		\begin{tabular}{c | c | c } 
			\hline
			Layer Type 			&  Layer Size(or value)	&  	input size\\

			\hline
			
			Dense + ReLU & 2000 & 1x1x784	\\ 			
			Dense + ReLU & 2000 & 1x1x2000	\\ 	
			Dense  & 10 & 1x1x2000	\\ 	
			Softmax & -  &  1x1x10		\\ \hline
		\end{tabular}
	}
	\label{table:PermutedMNIST}
\end{table*}

\begin{table*}[h]
	\centering
	\caption{Network Architecture for Split CIFAR-10/CIFAR-100 Experiment} \vspace{0.2cm}
	\scalebox{0.95}{
		\begin{tabular}{c | c | c } 
			\hline
			Layer Type 			&  Layer Size(or value)	&  	input size\\

			\hline
			
			Conv + ReLU & 3x3x4 & 32x32x3	\\ 			
			Conv + ReLU & 3x3x4 & 32x32x4	\\ 	
			Max Pooling & 2x2 & 32x32x4	\\ 
			Dropout & 0.25 & 16x16x4 \\
			Conv + ReLU & 3x3x8 & 16x16x4	\\ 
			Conv + ReLU & 3x3x8 & 16x16x8	\\ 
			Max Pooling & 2x2 & 16x16x8	\\ 
			Dropout & 0.25 & 8x8x8 \\
			Dense + ReLU  & 64 & 1x1x512	\\ 	
			Dropout & 0.5 & 1x1x64 \\
			Dense  & 90 & 1x1x64	\\ 	
			Softmax (Per-task) & -  &  1x1x90		\\ \hline
		\end{tabular}
	}
	\label{table:CIFAR}
\end{table*}

\begin{table*}[h]
	\centering
	\caption{Network Architecture for Tiny ImageNet Experiment} \vspace{0.2cm}
	\scalebox{0.95}{
		\begin{tabular}{c | c | c } 
			\hline
			Layer Type 			&  Layer Size(or value)	&  	input size\\

			\hline
			
			Conv + ReLU & 3x3x32 & 224x224x3	\\ 			
			Max Pooling & 2x2 & 224x224x32	\\ 
			Dropout & 0.25 & 112x112x32 \\
			Conv + ReLU & 3x3x32 & 112x112x32	\\ 
			Max Pooling & 2x2 &  112x112x32	\\ 
			Dropout & 0.25 & 56x56x32 \\
			Conv + ReLU & 3x3x64 & 56x56x32 	\\ 
			Max Pooling & 2x2 &   56x56x64	\\
			Dropout & 0.25 & 28x28x64 \\
			Conv + ReLU & 3x3x64 & 28x28x64 	\\ 
			Max Pooling & 2x2 &   28x28x64	\\
			Dropout & 0.25 & 14x14x64 \\
			Conv + ReLU & 3x3x64 & 14x14x64 	\\ 
			Max Pooling & 2x2 &   14x14x64	\\
			Dropout & 0.25 & 7x7x64 \\			
			Dense + ReLU  & 2048 & 1x1x3136	\\ 	
			Dropout & 0.5 & 1x1x2048 \\
			Dense  & 180 & 1x1x2048	\\ 	
			Softmax (Per-task) & -  &  1x1x180		\\ \hline
		\end{tabular}
	}
	\label{table:TinyImageNet}
\end{table*}

\end{document}